\documentclass[letterpaper]{article}
\usepackage{aaai19}  
\usepackage{times}  
\usepackage{helvet}  
\usepackage{courier}  
\usepackage[hyphens]{url} 
\usepackage{graphicx}  
\usepackage{placeins}

\usepackage{colortbl}
\usepackage{nth}
\usepackage{capt-of}
\usepackage{comment}
\usepackage{multirow}
\usepackage{floatrow}
\usepackage{float}
\usepackage{booktabs}
\usepackage{enumitem}
\usepackage{multicol}
\usepackage{graphicx}
\usepackage{subcaption}
\usepackage{amsmath}

\usepackage{arydshln}

\begin{document}
\title{NELA-GT-2022: A Large Multi-Labelled News Dataset for The Study of Misinformation in News Articles}

\author{Maur\'{i}cio Gruppi\textsuperscript{*}, Benjamin D. Horne\textsuperscript{\dag} and Sibel Adal{\i}\textsuperscript{*}\\
 Rensselaer Polytechnic Institute\textsuperscript{*}, The University of Tennessee Knoxville\textsuperscript{\dag}\\
 gouvem@rpi.edu, bhorne6@utk.edu, adalis@rpi.edu
}

\maketitle

\begin{abstract} 
In this paper, we present the fifth installment of the NELA-GT datasets, \texttt{NELA-GT-2022}. The dataset contains 1,778,361 articles from 361 outlets between January 1st, 2022 and December 31st, 2022. Just as in past releases of the dataset, \texttt{NELA-GT-2022} includes outlet-level veracity labels from Media Bias/Fact Check and tweets embedded in collected news articles. The NELA-GT-2022 dataset can be found at: \url{https://doi.org/10.7910/DVN/AMCV2H} 
\end{abstract}

\section{Introduction}
Many disciplines utilize news media data in their research, ranging from the study of mass media in journalism to the building of automated tools in computer science. Across these areas, researchers need historical article data that is consistent across time and covers many different types of news outlets. In specific studies of `fake news' detection,  large news datasets with veracity labels are needed.

To these many ends, researchers have focused on building high-quality news datasets. There are several platforms dedicated to collecting news data, such as Media Cloud, an open source platform used for collecting and analyzing global news coverage \cite{roberts2021media}, and LexisNexis, a commercial news database often used in academic studies \cite{deacon2007yesterday}. There are also many one-time news data collections. For example, the FA-KES dataset \cite{salem2019fa}, the Golbeck et al. dataset  \cite{golbeck2018fake}, and the Election-2016 dataset \cite{bode2020words,bozarth2020toward}. Other one-time datasets focus on social media posts rather than news articles, such as the FakeNewsNet dataset \cite{shu2018fakenewsnet}. 

While all of these data sources have been useful for a variety of research studies, there continues to be a need for updated news data. Platforms like Media Cloud do an excellent job at capturing high-quality, current news coverage around the world, but do not capture low-veracity news outlets. Datasets like the Golbeck et al. dataset and the FA-KES datset capture low-veracity news, but quickly become out-dated. The yearly-released NELA-GT datasets continue to fill both these gaps: updated news coverage across both low and high veracity outlets.

In this short paper, we describe the fifth release of the NELA-GT datasets, \texttt{NELA-GT-2022}. In \texttt{NELA-GT-2022} we have collected \textbf{1,778,361 articles} from \textbf{361 outlets} between \textbf{January 1st, 2022 and December 31st, 2022}. Included with these news articles are outlet-level veracity labels from Media Bias Fact Check, with \textbf{337 of 361 outlets labeled}, and data on \textbf{346,283 distinct tweets} embedded into collected news articles.

In this paper, we describe what is new in the 2022 version of the dataset, the collection process, the publicly-available data formats, and potential use cases.

\section{What's New in NELA-GT-2022?}~\label{sec:whatsnew} 
\begin{figure*}
    \centering
    \begin{subfigure}{0.6\textwidth}
        \includegraphics[width=\textwidth]{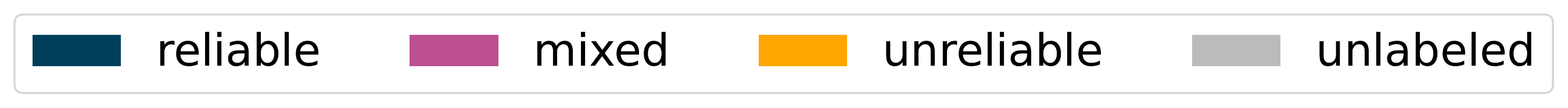}
    \end{subfigure} ~
    \begin{subfigure}{0.48\textwidth}
    \includegraphics[width=\textwidth]{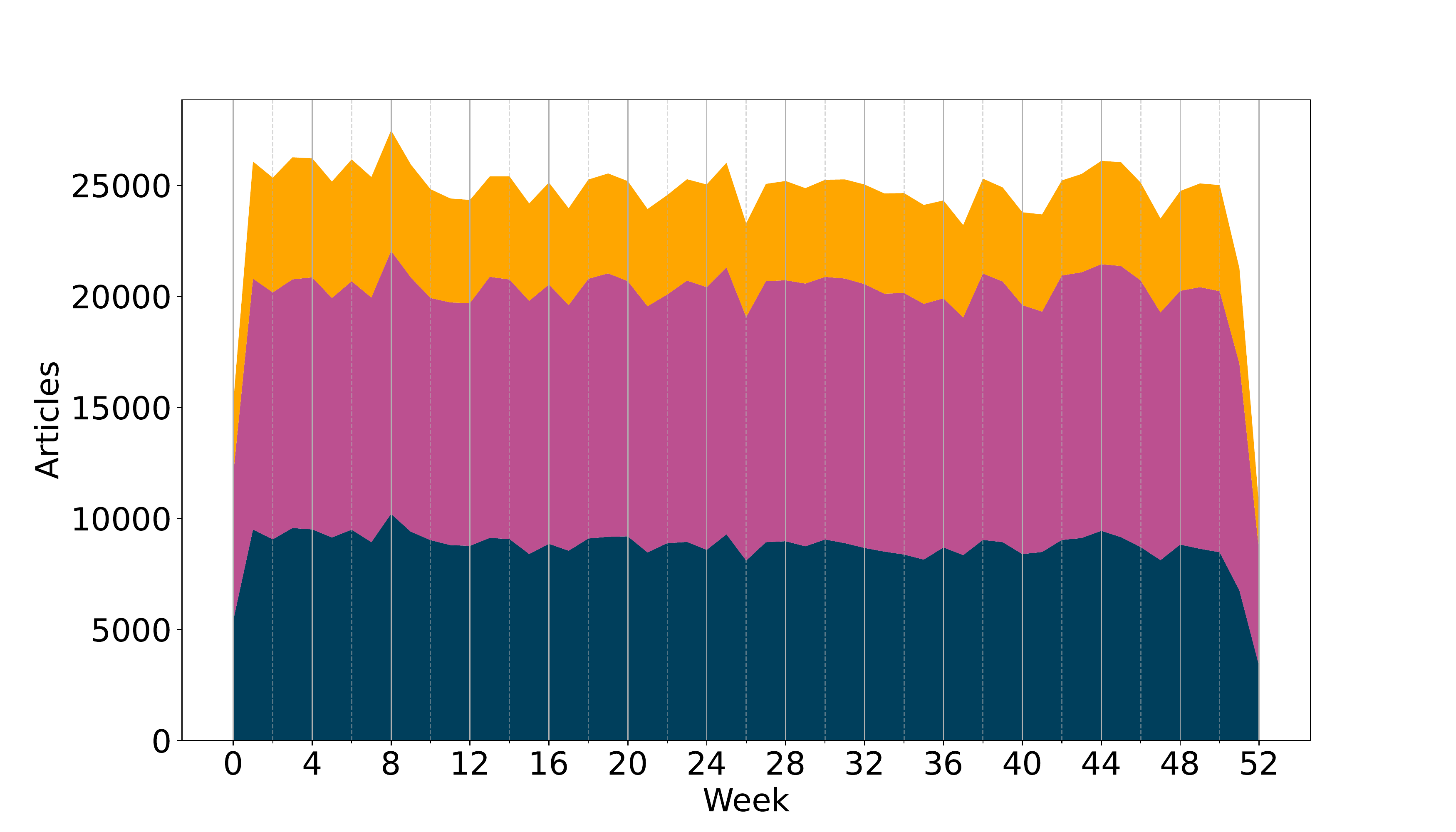}
    \caption{Number of articles per reliability class.}
    \end{subfigure} ~
    \begin{subfigure}{0.48\textwidth}
    \includegraphics[width=\textwidth]{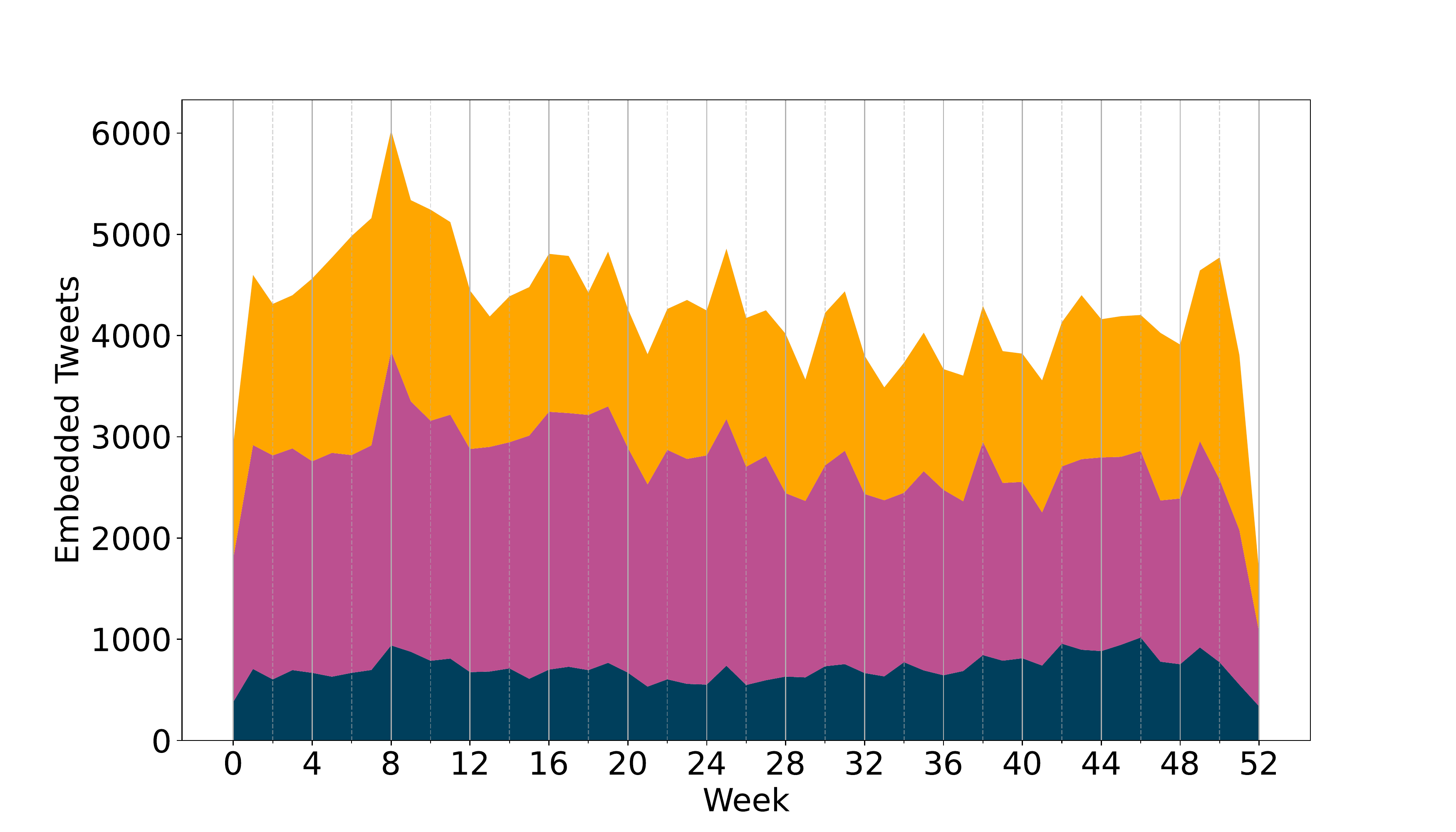}
    \caption{Number of embedded tweets per reliability class.}
    \end{subfigure}
    
    \begin{subfigure}{0.48\textwidth}
    \includegraphics[width=\textwidth]{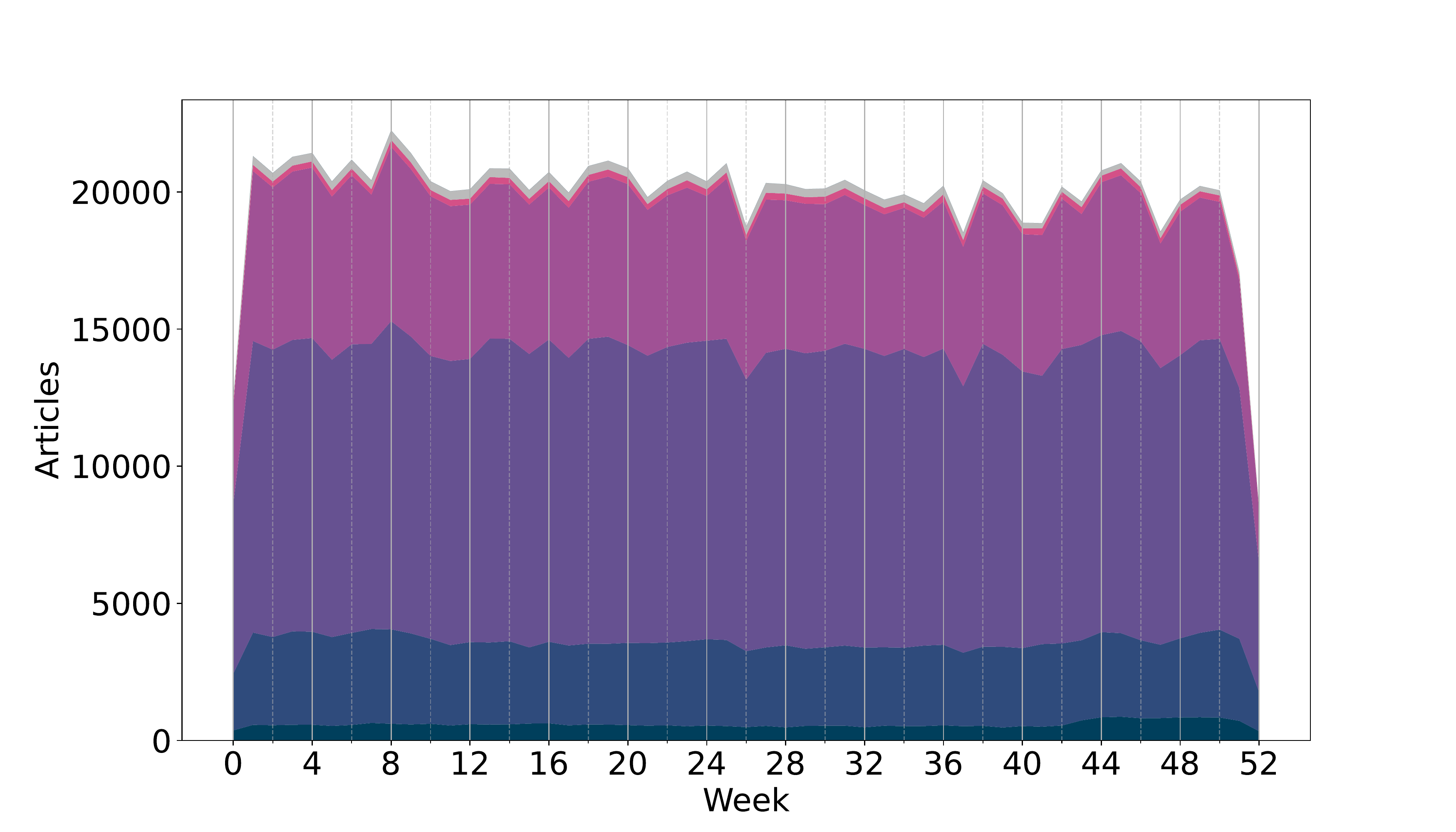}
    \caption{Number of articles per MBFC factuality score.}
    \end{subfigure}~
    \begin{subfigure}{0.48\textwidth}
    \includegraphics[width=\textwidth]{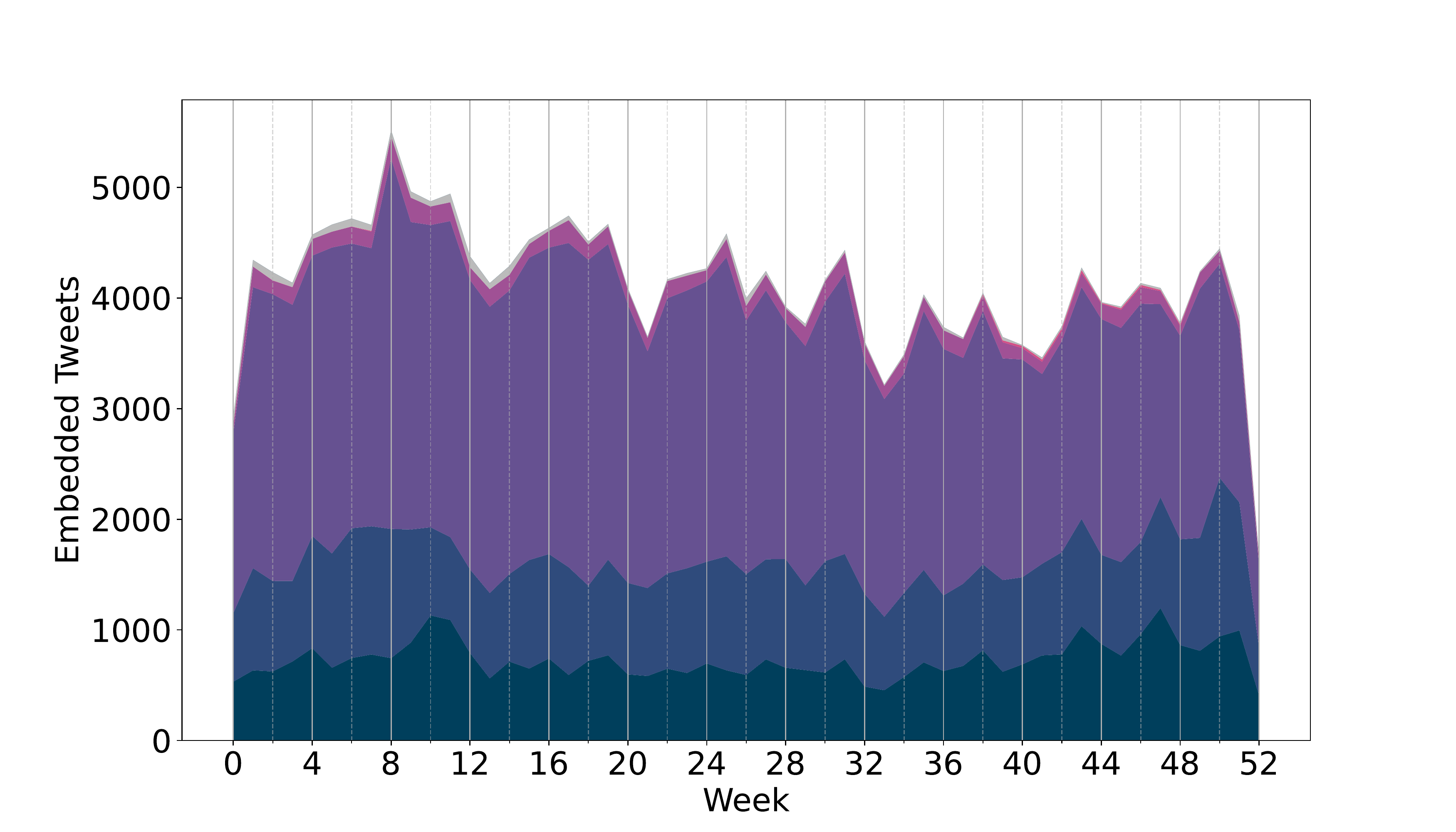}
    \caption{Number of embedded tweets per MBFC factuality score.}
    \end{subfigure}
    
    \begin{subfigure}{0.6\textwidth}
        \includegraphics[width=\textwidth]{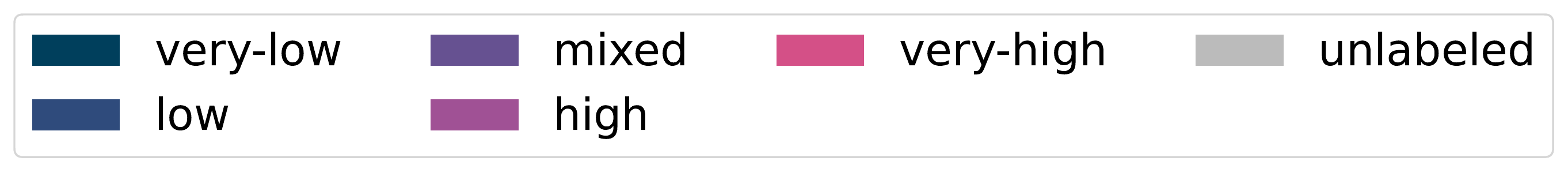}
    \end{subfigure}
    \caption{Number of articles (a, c) and embedded tweets (b, d) collected during each week of 2021.}
    \label{fig:data-weekly}
\end{figure*}

Again, our focus this year was to stabilize our collection infrastructure to ensure complete coverage of articles published across the full year, rather than add new features to the dataset. Hence, as shown in Figure \ref{fig:data-weekly}, we estimate that our collection has little to no missing article data in 2022.


\begin{figure*}
    \centering
    \begin{subfigure}{0.45\textwidth}
    \includegraphics[width=\textwidth]{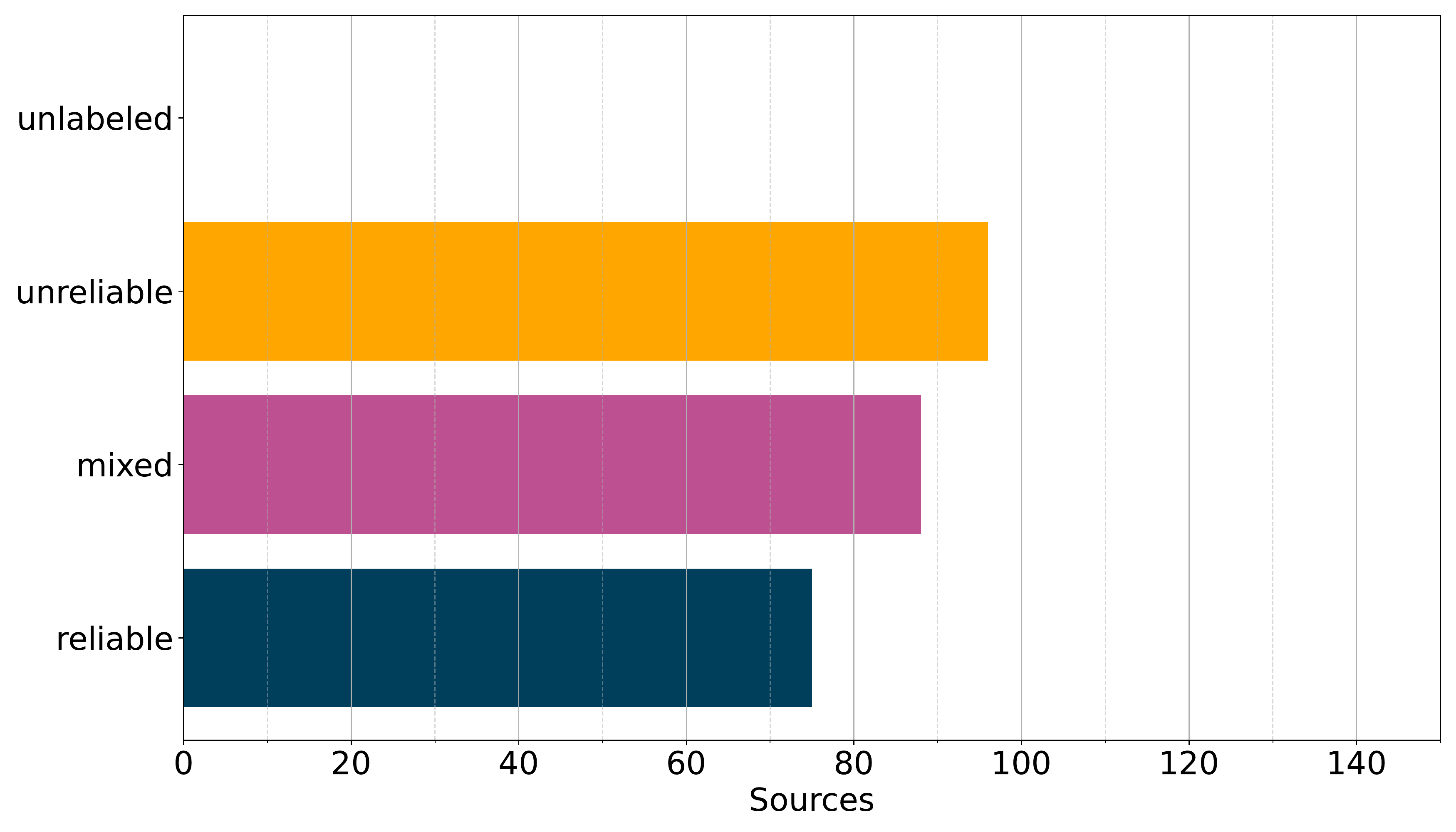}
    \caption{Number of sources per reliability class.}
    \end{subfigure} \qquad
    \begin{subfigure}{0.45\textwidth}
    \includegraphics[width=\textwidth]{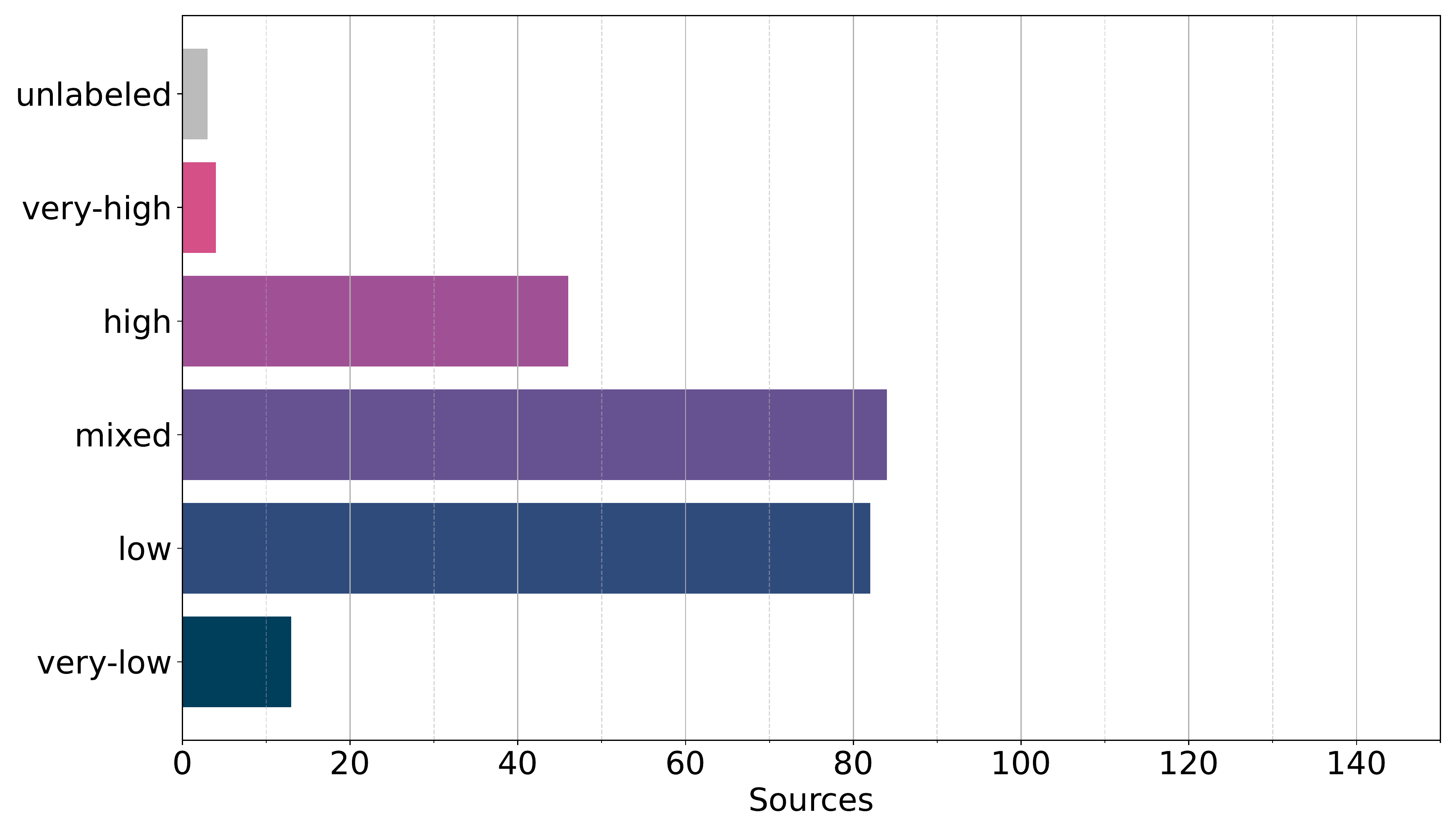}
    \caption{Number of sources per MBFC factuality score.}
    \end{subfigure}
    
    \caption{Distribution of sources per reliability class (a) and factuality (b) score.}
    \label{fig:sources-per-category}
\end{figure*}

\section{Data Collection}

\subsection{News data and metadata}
The data collection process follows what was described in \cite{norregaard2019nela}.  Specifically, we scraped the RSS feeds of each outlet in our outlet list twice a day starting on 01/01/2022 using the Python libraries feedparser and Goose3\footnote{\url{https://github.com/grangier/python-goose}}. This list of outlets to collect was carried over from \cite{gruppi2021nela}. These sources come from a variety of countries (or the country of origin is not known), but are all articles are in English.

See Table \ref{tab:format-newsdata} for details on each attribute stored during the collection process.

\subsection{Embedded tweet data}
In 2020, we introduced additional data on tweets embedded into news articles \cite{gruppi2020nela}. We again collect this data for the 2022 dataset. Specifically, we collected embedded tweets on the article page using the Goose3 library. The ID of the embedded tweet is stored in the database table tweet, along with the id of the article from which it was
collected and the text of the tweet. We show this structure in Table \ref{tab:format-tweet}.

\subsection{Limitations}
Since the articles collected from news sources may be copyrighted, 
we apply a transformation to the original text so that it cannot be 
used for their originally intended purpose, i.e., that of being
read by individuals to consume journalistic information. 

Specifically, we modify the text so that it cannot properly be used for news
consumption but can still be used for text analysis by periodically removing words from the text. For articles with more than 200 tokens, we replace 7 tokens with `@` every 100 tokens. For articles with fewer than 200 tokens, we replace 5 consecutive tokens with `@` every 20 tokens.
This transforms the articles so that it is unlikely that a user will
read NELA-GT to consume news while still keeping most of the content
that is useful for analysis (approximately 7\% for larger articles).

\begin{table*}[ht!]
    \centering
    \begin{tabular}{c|c|l}
         \textbf{Column} & \textbf{Type}  & \textbf{Description}\\ \toprule
         id & text (primary key) & Article identifier. \\
         date & text & Publication date string in YYYY-MM-DD format. \\
         source & text & Name of the source from which the article was collected. \\
         title & text & Headline of the article. \\
         content & text & Body text of the article. \\
         author & text & Author of the article (if available). \\
         published & text & Publication date time string as provided by source (inconsistent formatting). \\
         published\_utc & integer & Publication time as unix time stamp. \\
         collection\_utc & integer & Collection time as unix time stamp. \\
         url & Text & URL of the article.
    \end{tabular}
    \caption{Structure of NELA-GT-2022 article data. For the database format, column \textbf{id} is the primary key of table \texttt{newsdata}.}
    \label{tab:format-newsdata}
\end{table*}

\begin{table*}[ht!]
    \centering
    \begin{tabular}{c|c|l}
         \textbf{Column} & \textbf{Type}  & \textbf{Description}\\ \toprule
         id & text (primary key) & Tweet id. \\
         article\_id & text (foreign key) & Id of the article in which the embedded tweet was observed. \\
         embedded\_tweet & text & ID/URL of the embedded tweet.
    \end{tabular}
    \caption{Structure of NELA-GT-2022 embedded tweets. For the database format, column \textbf{id} is the primary key of table \texttt{tweet}.}
    \label{tab:format-tweet}
\end{table*}

\section{Format of Data}\label{format}
Just as in the past three versions (2019, 2020, 2021), the dataset has been released in two formats: (1) a SQLite database, (2) a JSON dictionary per news source. Details about the structure of each of these formats is below. We provide Python code to read both data formats at: \url{https://github.com/MELALab/nela-gt}.

The SQLite 3 database schema consists of two tables: \texttt{newsdata} and \texttt{tweet}. The \texttt{newsdata} table contains, in each row, data about an article. Column \textbf{id} is set as primary key to avoid duplicated entries on the database. We normalized source names by converting them to lower case, and removing spaces, punctuation, and hyphens. For example, the source \emph{The New York Times} appears as \emph{thenewyorktimes},  Tables \ref{tab:format-newsdata} and \ref{tab:format-tweet} give information about data columns.

\subsection{JSON Format}\label{json}
We also provide the dataset in JSON format. Specifically, each source has one JSON file containing the list of all of its articles. The fields follow the same structure of the database columns (Tables \ref{tab:format-newsdata} and \ref{tab:format-tweet}).

\subsection{Ground Truth Data Format}
We include multiple types of source-level veracity labels. In \texttt{NELA-GT-2022}, we collect source-level labels from Media Bias/Fact Check (MBFC) that contain the following dimensions of veracity:
\begin{enumerate}
    \item Media Bias Fact Check factuality score - on a scale from 0 to 5 (low to high credibility).
    \item Media Bias Fact Check Conspiracy/Pseudoscience and questionable sources - low credibility if a source belongs to these categories.
\end{enumerate}

In addition, we create an aggregated version of the factuality scores, broken down into three classes: \textit{reliable}, \textit{mixed}, and \textit{unreliable}.

Due to the limited availability of veracity labels from other platforms, we choose to only collect labels from MBFC. However, we encourage researchers to use and compare veracity labels from multiple resources when possible. This is particularly important when testing machine learning models. For an overview of the impact of ground truth labels on news studies, please see \cite{bozarth2020higher}. Furthermore, we strongly encourage machine learning researchers to test news veracity models using robust evaluation frameworks, such as those discussed in \cite{bozarth2020toward} and \cite{horne2019robust}.

\section{Use Cases}
\subsection{Analysis of news coverage during events}
One of the primary goals in the yearly-release of the NELA-GT datasets is to provide updated coverage of current events. To this end, we provide two example subsets of the database for two events during 2022: the Russo-Ukrainian War and the overturning of Roe v. Wade.

\subsubsection{Russo-Ukrainian War}
In February 2022, Russia launched an invasion of Ukraine, connecting back to conflict starting in 2014. This event came to many as a surprise, as Russian officials repeatedly denied plans to attack Ukraine. his international event had many ramifications across the globe, including gas pipeline disputes and global disinformation campaigns about the war\footnote{\url{https://en.wikipedia.org/wiki/Russo-Ukrainian_War}}. This event and its effects have been widely covered in media, including fringe, conspiracy media. 

\subsubsection{Overturning of Roe v. Wade}
Another major event in 2022 was the overturning of Roe v. Wade in the United States, which guaranteed the right to abortion for nearly 50 years. With Roe v. Wade overturned, federal standards on abortion access were removed, allowing many states to ban abortions and forcing many women's clinics to close\footnote{\url{tinyurl.com/3v5h96sr}}. Given the partisan divide on abortion rights in the United States, news coverage of this event covers a range of positions across political lines.

\begin{figure*}[ht]
    \centering
    \begin{subfigure}{0.45\textwidth}
    \centering
    \includegraphics[width=\textwidth]{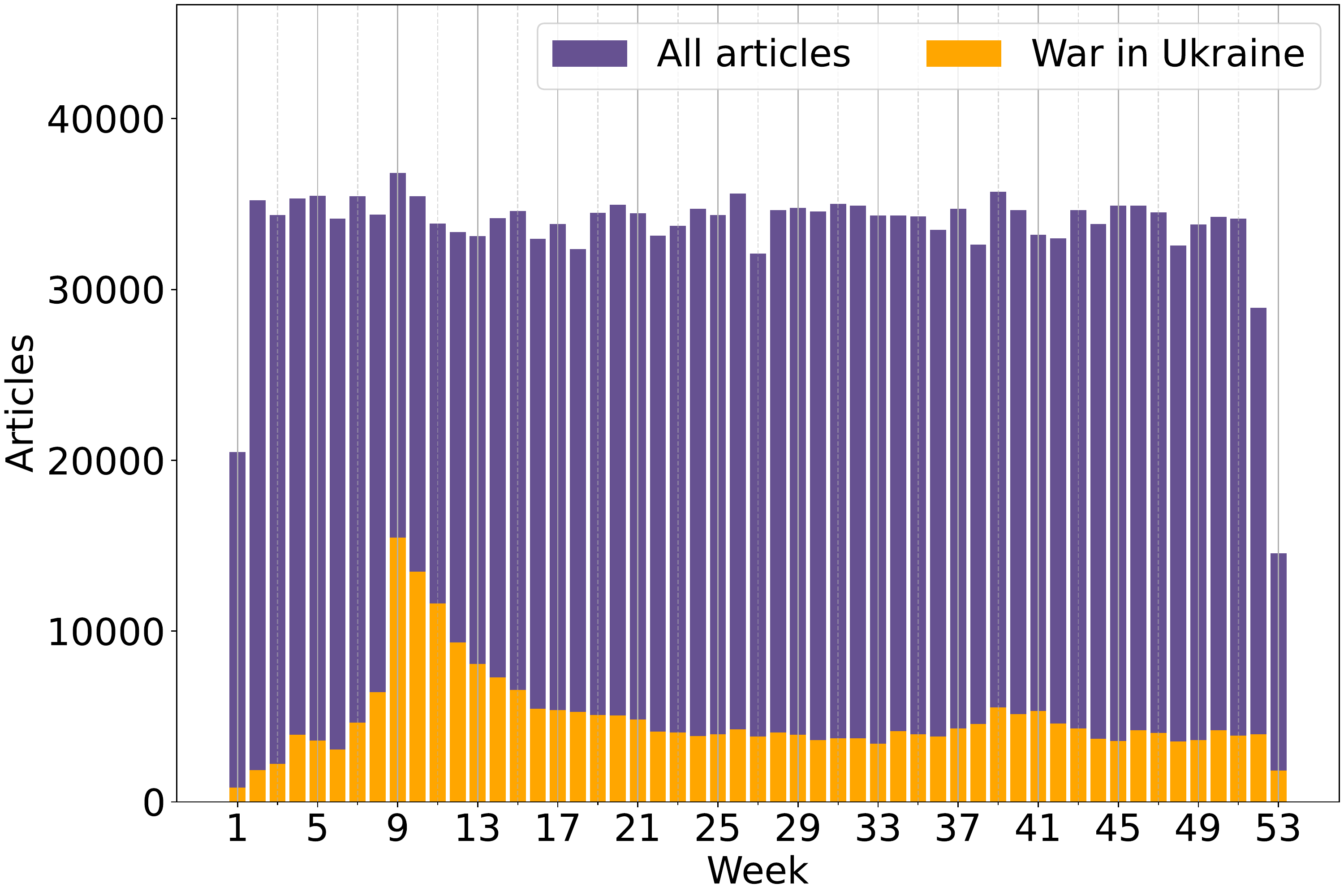}
    \caption{Number articles related to the Russo-Ukrainian War in comparison to all articles collected in NELA-GT-2022 in each week of 2022.}
    \label{fig:all-covid}    
    \end{subfigure} \qquad
    \begin{subfigure}{0.45\textwidth}
        \centering
        \includegraphics[width=\textwidth]{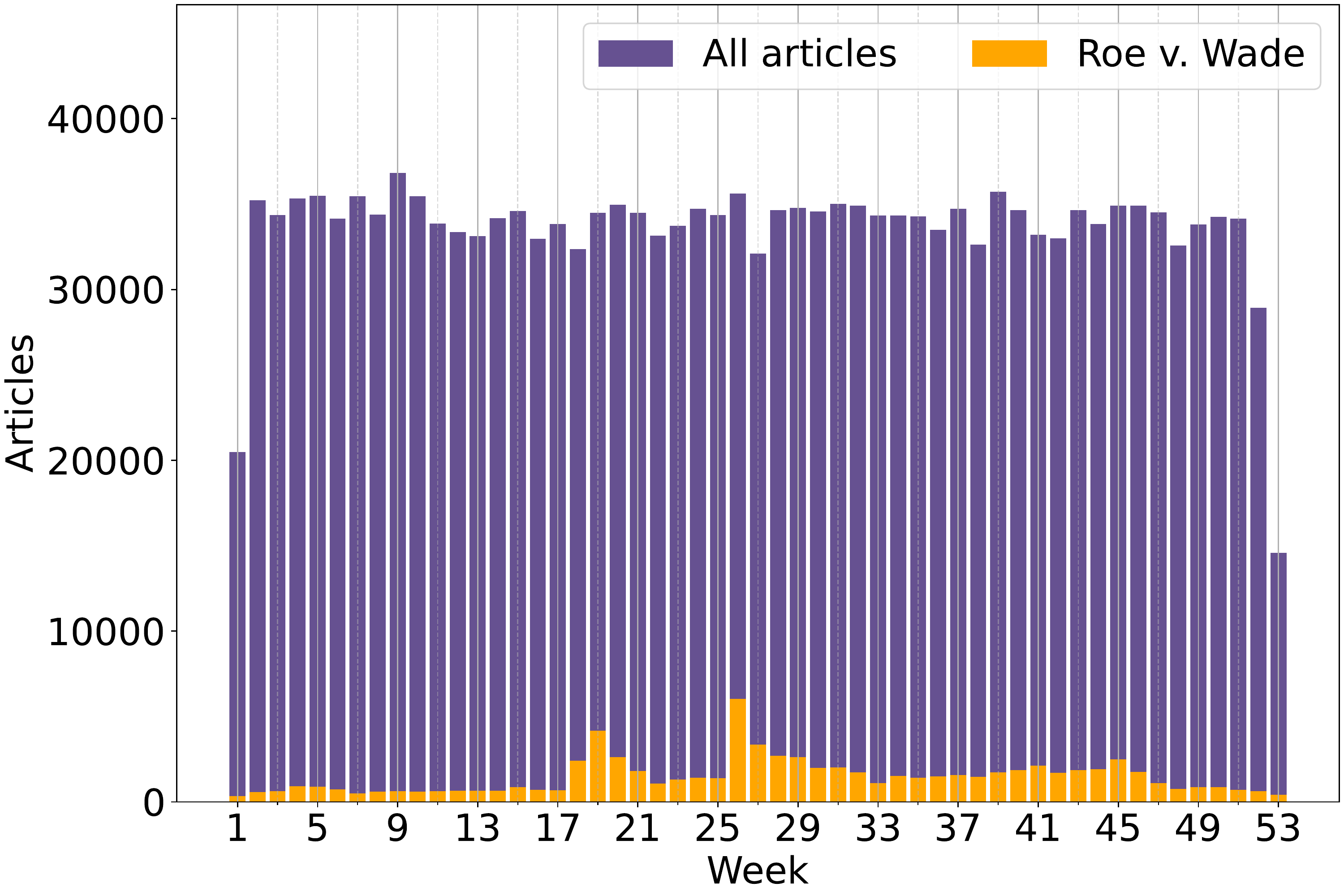}
        \caption{Number of articles related to overturning of the Roe v. Wade case in comparison to all articles collected in NELA-GT-2022 in each week of 2022.}
        \label{fig:all-election}
    \end{subfigure}
    \caption{Number of articles related to (a) Russo-Ukrainian War, and number of articles related to (b) the overturning of the Roe v. Wade case as a fraction of the total number of articles in each week of 2022. Articles are found using a set of keywords shown in \ref{tab:keywords}.}
    \label{fig:all-articles}.
\end{figure*}

Figure \ref{fig:all-covid} and \ref{fig:all-election} show the number of article related to these events over time in the dataset.

\subsection{Embedded Tweets}
By providing the tweets embedded in news articles, the \texttt{NELA-GT-2022} dataset can be useful furthering in studies of political communications and hybrid media systems. Notably, very few studies have addressed low-veracity news sources role in these hybrid systems, which this dataset can aid \cite{gruppi2021tweeting}. 

\subsection{Long-Term Use Cases}
Our primary goal with the continued release of the NELA-GT datasets is to support long-term news research. When combining all of the NELA datasets (both the NELA-GT datasets \cite{norregaard2019nela,gruppi2020nela,gruppi2021nela,gruppi2022nela} and the original NELA2017 dataset \cite{horne2018sampling}), we provide over 6.1M news articles across 5.5 years. There are multiple research avenues that this data, both in part and as a whole, supports: 
\begin{itemize}
    \item Exploring event-driven dynamics of and narratives in news media: Analyses of narrative themes before, during, and after major events continues to be a useful methodology in interdisciplinary media studies. This dataset supports these works by maintaining consistent data collection across events. 
    \item Robust machine learning: This dataset allows for continued work in automated news veracity detection, particularly in robustness checks of current work. These robustness checks include examining prediction accuracy over time, over events, and over mixed veracity labels. We again encourage machine learning researchers to test news veracity models using robust evaluation frameworks, such as those discussed in \cite{bozarth2020toward}, and to use multiple datasets when possible.
    \item Examining media manipulation: Using the veracity labels in this dataset, research can examine tactics used by hyper-partisan news outlets. Additionally, with knowledge of media manipulation campaigns, such as those discussed in the Media Manipulation Casebook\footnote{\url{https://mediamanipulation.org/}}, researchers can examine how media manipulation is propagated through malicious news outlets. While there has been a substantial focus on “fake news” detection methods by researchers, there still is room to better understand media manipulation and disinformation campaigns.
\end{itemize}

\section{Conclusion}
In this paper, we describe the \texttt{NELA-GT-2022}, a dataset of news articles from sources of varying veracity.
The RSS feeds from the sources were scraped twice a day on every day of 2022, resulting in a set with 1,778,361 articles from 361 outlets. The dataset includes the source factuality labels from Media Bias Fact Check and tweets that were embedded in the collected news articles. We provide two event-based subsets of the dataset for the study of news coverage and messaging around the war in Ukraine and the overturning of Roe v. Wade. These subsets were generated from the original dataset by filtering articles based on keyword matching.

The dataset and additional documentation can be found at: \url{}. Example code for data extraction can be found at: \url{https://github.com/MELALab/nela-gt}. 

\begin{table}[ht!]
    \centering
    \begin{tabular}{c|c}
     \textbf{Russo-Ukrainian War keywords} & \textbf{Roe v. Wade keywords}\\ \toprule
        Ukraine & roe\\
        Russia & wade\\
        Ukraine war & abortion\\
        Ukraine and Russia & abortions\\
        Ukrainian service members & pro-life\\
        Ukrainian servicemen & pro-choice\\
        Russian soldiers & overturned\\
        Russian troops & abortion ban\\
        Russian forces & anti-abortion\\
        Russian-backed forces & planned parenthood\\
        Ukrainian military\\
        Donetsk\\
        Luhansk\\
        Donbas\\
        Lyman\\
        Lysychansk\\
        Bakhmut\\
    \end{tabular}
    \caption{Keywords used to make the event-based data subsets for the War in Ukraine and Inflation. The full keyword lists are provided with the dataset.}
    \label{tab:keywords}
\end{table}

\bibliographystyle{aaai}
\bibliography{main}
\end{document}